# Machine Learning Algorithms for Depression Detection and Their Comparison.

Danish Muzafar[1], Furqan Yaqub Khan[2], Mubashir Qayoom[3]
[1]Department of Information Technology, Central University of Kashmir, J&K, India
[2]Department of CSE, IIT Patna
[3]Department of Physics, University of Kashmir, J&K, India
Email: furkaan309@gmail.com

***Abstract*** *–Textual emotional intelligence is playing a ubiquitously important role in leveraging human emotions on social media platforms. Social media platforms are privileged with emotional contents and are leveraged for various purposes like opinion mining, emotion mining, and sentiment analysis. This data analysis is also levered for prevention of online bullying, suicide prevention and depression detection among social media users. In this article we have designed an automatic depression detection of online social media users by analyzing their social media behavior. The designed depression detection classification can be effectively used to mine user's social media interactions and one can determine whether a social media user is suffering from depression or not. The underlying classifier is made using state-of-art technology in emotional artificial intelligence which includes LSTM (Long Short Term Memory) and other machine learning classifiers. The highest accuracy of the classifier is around 70% of LSTM and for SVM the highest accuracy is 81.79%. We trained the classifier on the datasets that are widely used in literature for the emotion mining tasks. Confusion matrix of results is also given.*

***Keyword-***--Emotion Artificial Intelligence, Deep Learning, Machine Learning Algorithms, Word Embedding's, Twitter.

## I. INTRODUCTION

Depression triggers sorrow, frustration, lack of interest, annoyance. It will contribute to variety of emotional and physical problems and might cut back the power of a person to operate at full potential [1]. The matter of mental state in humans is incredibly serious. At its worst, depression is the main cause of death for 80000 human beings leading them to suicide. Teenagers are most prone to depression as a result of they often need Psychotherapy,

Teenagers limitlessly use mobile phones and social media and spend less time in real world and more in virtuality which causes a gap between their internal state of mind and outward world, causing a strong sense of loneliness which in-turn leads to depression[2]. Bullying via social media based on factors like social status, standard of language, race, ethnicity, immigration standards, health, behavioral ethics, etc. are the main drawbacks of social media which cause depression to user.[4] The definition of problem dealt in this research article is to find whether the person under consideration does have symptoms of depression or not. With the use of neural networks we have used users tweets to diagnose whether the person under consideration is behaving positively or negatively. If the person is shows particular standard of positivity in his tweets then we can predict him or her to be depression free, otherwise he or she is suffering from depression. The standard used in this article for a person to be considered under depression is that if 80% of his tweets are classified negative. We do the analysis of user feeds posted on a twitter account via state-of-art machine learning and deep learning techniques based on scientifically accepted 80% criterion.[6][7].

1.1. Some Initial Symptoms of depression are listed below:

- Feeling crazy, sad or in a depressing mood
- Weight loss
- Increase fatigue and sleep difficulties.
- Loss of energy and increased fatigue with urge to suicide.
- Thinking very difficultly, concentrating, or making decisions based on the assumption based on the idea of not liking anything.
- Frequent and continuous of death or suicide

Symptoms must last at least two weeks or longer than that. Depression may have a different effect on the sexes and shows that men with depression can have signs such as irritability, escape or dangerous behavior, or that everyone is really angry. Symptoms must last at least two weeks or even more for a diagnosis of depression [1]

1.2. Risk Factors for depression- Depression can affect anyone, even a person who seems to be living in reasonably ideal conditions.

Depression can be caused by a variety of circumstances:

- *Biochemistry:* Harmonal differences may contribute to symptoms of depression in certain cases due to chemicals imbalance in the brain.

- *Genetics:* Genetics also plays an essential role in co-depression for e.g. if one identical twin has depression, for instance, the other has a 70 percent risk of developing the disorder sometime in life.
- *Personality:* People with low self-confidence are more likely to experience depression, causing them to be speechless by stress, or who are generally pessimistic.
- *Environmental factors:* Continuous exposure to violence, negligence, abuse or poverty can make some individuals more susceptible to depression.

*1.3. Role of Social Media and Depression*

Social networks are becoming an important part of the lives of individuals, they mirror the personal life of the user on social media, people like to share happiness, joy and sorrow [4]. Researchers are using these platforms to classify and detect the causes of depression, reading an old article on the News website about how Twitter can tell if you're depressed and the possibility of developing an artificial intelligence model that can scan your Twitter feed and tell you if an individual is at risk of depression, or accepting notifications from third parties, such as those advising you to seek help based on an automatic scan of your tweets [1]. The day has come to an end. Early detection of depression can be a big step toward addressing the mental illness and providing help to those who are suffering from it. Various approaches for sentiment analysis in Machine Learning, including decision-based systems, Bayesian classifiers, support vector machines, neural networks, and sample-based methods can be used to detect symptoms of depression before it is too late.

**II. Depression and Emotion Intelligence**

If a person is attached with someone closely then it is clear the person is emotionally attached to that person and if this emotional dependence is too high that the decisions of attached persons are dependent on each other this is called emotion bonding. This is when these persons take emotions seriously sometimes their emotions are not positive in nature hence leading to depression in one or all of them.

**2.1. Emotional intelligence is commonly defined by four attributes:**

a. **Self-management** – You can control impulsive emotions and actions, handle your emotions in healthy ways, take action, meet commitments, and respond to changing conditions.

b. **Self-awareness** – You recognize your own feelings and how your thoughts and conduct are affected by surrounding conditions. You are aware of your strengths and weaknesses, and have faith and confidence in ones self.

   c. **Social awareness** – You've got empathy. You should understand other people's thoughts, desires, and concerns, pick up on emotional signals, socially feel relaxed, and know the dynamics of power in a group or organization.

   d. **Relationship management** – You know how to build and maintain good relationships, clearly communicate, inspire and influence others, work well as a team, and resolve conflicts.

## III. DISCUSSION

### 3.1. Dataset

We have developed a depression detection dataset by manually annotating the twitter data. We used two classes for this research one is depressive and other non-depressive. The Tweets were downloads from the twitter Platform and some from the kaggle data and initial dataset had 25,000 tweets from which foreign language tweets and re-tweets were removed. This dataset contains three columns: First Column contains tweet ID, second column contains text and third column contain labels these columns are features. The dataset was then annotated by three annotators and majority voting was kept. Thus if a tweet was annotated depressive by more than one annotator only then it was given a depressive label likewise was done for non-depressive tweets. In the final dataset we had 7,500 depressive tweets and 7,500 non-depressive tweets.

Different algorithms are used we discuss one by one:

### 3.2. Support Vector Machine (SVM)

Support Vector Machine(SVM) is a supervised algorithm for machine learning that can be used for problems with classification and regression. In classification issues, however, it is mostly used [1]. In the SVM algorithm, we depict each data item as a point in n-dimensional space (where n is the number of characteristics you have), with the value of each characteristic being the coordinate value. Then, we carry out classification by determining the hyper-plane that clearly distinguishes the two classes [12]. The Support Vector Machine is a discriminative classifier that is formally described by a distinct hyperplane (SVM). In other words, given labelled training data, the algorithm generates an ideal hyperplane that categorizes fresh examples (supervised learning). This hyperplane is a line in two-dimensional space that divides a plane into two sections, with each class on either side. The learning of the hyperplane in linear SVM is finished by changing the issue using some linear algebra [10]. This is when the kernel comes into play. In our experiment we gave the linear SVM using train test split of 80-20%. We employed scikit-learn kernels for the classification purpose.

### 3.3. Random Forest Classifier

The Random Forest Classifier an ensemble algorithm, produces a set of decision trees from the training set's randomly selected subset. To evaluate the final class of the test object, it then aggregates the votes from various decision trees. Different decision trees can create subsets that overlapping in nature. To deal with this weighting principle is used to consider the impact of any decision tree's result. Trees with a high mistake rate are given a low weight value, whereas trees with a low mistake rate are given a high weight value. Low error-rate trees would have a greater impact on decision-making as a result of this. The total number of trees to be constructed, are as well the important characteristics for the decision tree. The minimum split, split criteria, etc., are all basic Random Forest Classifier parameters. We used Random Forest classifier with 40 estimators. The training and testing sets as inputs to classifier were divided into 80-20 train test splits of total data entries. We employed scikit-learn kernels for the classification purpose as well.

### 3.4. Multinomial Naïve Bayes

Naive Bayes is a family of algorithms based on applying the Bayes theorem with the strong (naive) assumption that each feature is independent of the others in order to predict the category of a given sample. Because they are probabilistic classifiers, the Bayes theorem will be used to compute the likelihood of each category, and the category with the highest probability will be created. Naive Bayes classifiers have been used successfully in a variety of fields, including Natural Language Processing (NLP).Other options, such as Support Vector Machines (SVM) and neural networks, are available when dealing with NLP issues.

Their simplicity of Naive Bayes classifiers, on the other hand, makes them particularly desirable for such classifiers. Furthermore, they have been proved to be fast, reliable, and exact in a variety of NLP applications. We use a non-naive Bayes approach to look at sentences in their whole, thus if a sentence does not appear in the training set, we will get a zero probability, making further computations impossible. However, in the case of Naive Bayes, each word is assumed to be independent of the others [10]. We now examine individual words as a phrase rather than the complete sentence.

**3.5 Long Short Term Memory (LSTM)**

LSTMs have proven to be effective in tasks involving sequential knowledge. The input vectors used for the other classifiers, such as the TF-IDF weights vector, do not, however, maintain any information on the sequential relationship between each document's terms and phrases. Thus, the more appropriate input for LSTMs are sequences of word embedding vectors; in fact, word embedding are the preferred option for textual representation in modern deep learning.

We tuned the number of memory units, number of epochs, batch size, and input and recurring dropout rates of LSTM by using the training and tuning set. The number of epochs is the number of times in the neural network the entire training set is moved forward and backward [12]. The network can be fit and the resulting neuron weights may be far from optimal if the number of epochs is too low. Before each change to the weights, the batch size is simply the number of samples fed into the network and helps to make training more effective by reducing the memory requirements and the number of iterations needed to achieve the optimum weights. Finally, in an attempt to minimize overfitting, the dropout rate is simply the likelihood that an input or recurrent node will be omitted from consideration during a weight update. Input to the classifier in the 80-20 train test format was given to the training and testing set respectively.

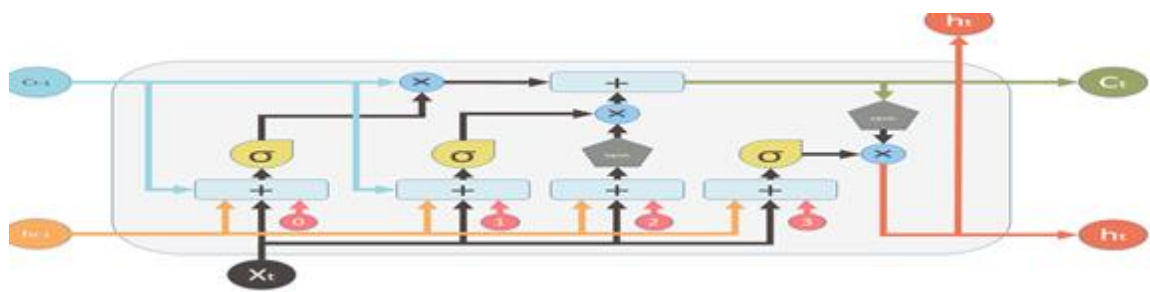

Fig. 1: Mechanism of LSTM Classifier

A cell, an input gate, an output gate, and a forget gate make up a standard LSTM unit. The cell recalls values across arbitrary time intervals, and the three gates govern the flow of information into and out of the cell. Because there might be lags of undetermined duration between critical occurrences in a time series, LSTM networks are well suited to categorizing, processing, and making predictions based on time series data. LSTMs were created to deal with the vanishing gradient problem that might occur when training regular RNNs. The relative insensitivity to gap length of LSTM over RNNs, hidden Markov models, and other sequence learning approaches provides a benefit a variety of applications.

**A. LSTM architecture:** Multiple architectures of LSTM units exist. The standard architecture consists of a cell (the LSTM unit's memory portion) and three information flow regulators inside the LSTM unit, typically referred to as gates: an input gate, an output gate, and a forgotten gate. Any of the versions of the LSTM device may not have one or both of these gates, or even other gates. Gated Recurrent Units (GRUs) do not, for example, have an output gate. The input gate determines how much a new value enters the cell, the forgotten gate determines how much the cell value remains, and the output gate determines how much the cell value is used to generate the LSTM unit's output activation. The activation function of the LSTM gates is also the logistic sigmoid function.

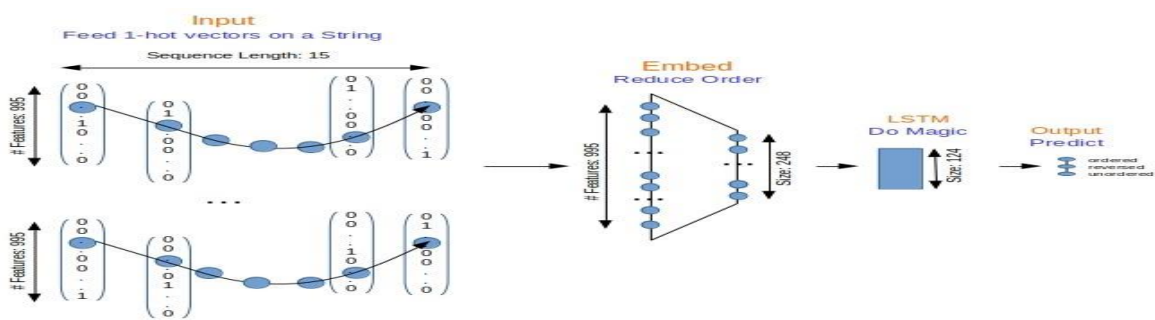

Fig. 2: Process of LSTM

**B. Word Embedding**

A word embedding is a group of ways for representing words and texts that use a dense vector representation. Instead, in an embedding, words are represented by dense vectors, each of which reflects the word's projection into a continuous vector space [2]. The placement of a word is based on the words that surround it when it is used and is learned from text inside the vector space. The embedding of a word is its position in the learned vector space. Two common examples of learning methods for embedding words from text include:

- Word2Vec.
- GloVe.

In this Research article Word2Vec and Doc2Vec are used.

**C. Keras Embedding layer**

Keras provides an embedding layer which can be used on text data for neural networks. This includes integer encoding of the input data, so that each word is represented by a unique integer. The Tokenizer API that is also supplied with Keras can be used to perform this data preparation stage. All of the words in the training dataset will learn an embedding since the Embedding layer is initialized with random weights. It's a versatile layer that can be applied in a variety of ways, like it can be used on its own to teach a word integration that can then be stored and reused in a different model. It can be utilized as part of a deep learning model where the embedding is taught alongside the model. It is a type of transfer learning, and can be used to load a pre-trained word embedding model. Weights are learned in the Embedding layer. If you save your model as a file, the weights for the Embedding layer will be included. An embedding layer's output is a 2D vector in the input word sequence (input document), with one embedding for each word. You must first flatten the 2D output matrix to a 1D vector if you want to link a dense layer directly to an embedded layer using the *Flatten* layer. Finally we train the LSTM Classifier for training and testing data classification.

**IV. METHOLODOLOGY**

In this section we discuss the overall methodology that is used for depression detection using online social profiles of the user. We employ emotional artificial intelligence for the depression detection task. We create a depression classifier using Deep-Learning framework and machine learning and use that classifier for classifying tweets from an online social profile for automatic depression detection. Figure 5 shows the overall methodology used for the task.

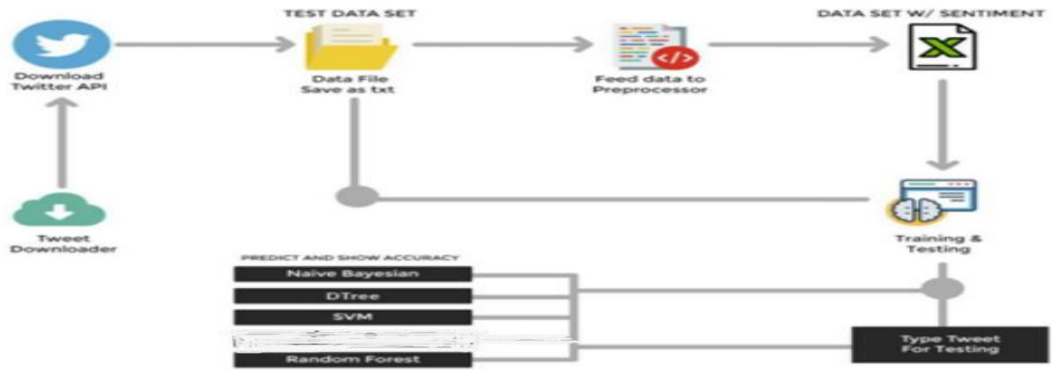

Fig. 3: Overall methodology of the depression detection system.

The system receives as input an emotion labeled dataset which is used for the training of supervised deep-learning classifier or machine learning classifier. Once the classifier is trained on emotion labeled dataset we use the tweets from online social profiles for the depression detection. If the tweets from the profile contain more than 75% tweets classified as depressive, our system classifies gives the red flag for the profile that it may be having depression symptoms. Figure 6 shows the components of the system in detail.

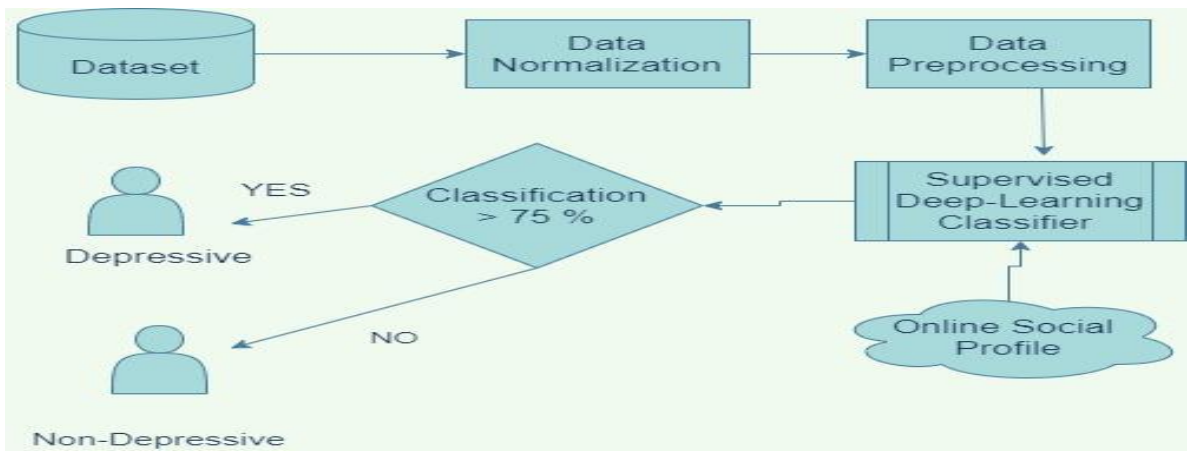

Fig. 4: Architecture of the system

The system has 5 modules and uses emotion labeled dataset.

### 4.1. Data Normalization

As part of data preparation for machine learning, standardization is a technique sometimes used. The aim of adjustment is to adjust the numeric column values in the dataset to use a common scale, without distorting disparities in the value ranges or losing information. You also deleted all retweets, all non-alphanumeric characters, URLs, and @mentions. Besides, all stop words were abolished, with the exception of first, second and third person pronouns. About 40% work done by this through NLTK automatically and all libraries.

**4.2. Data Preprocessing**

Preprocessing each tweet before passing it to the automated classifier is how the preprocessing module prepares it for the classifier. Slangs and abbreviations are eliminated by the use of English directories from the tweets. If a term is identified that does not have a meaningful definition, all of the words are passed into the dictionary module to be looked up, and then into the word replacer module to be replaced with the right word. In the word replacer module, we employed the SMS dictionary, Netlingo, and the urban dictionary. Preprocessing stages are made up of the following steps:

**4.3. Tokenization**

Is performed using Stanford Core NLP package to break tweets into sentences and words, such as units of words, sentences, or themes? The first column of the csv file containing the tweet is extracted and converted to individual tokens in this case.

**4.4. Stemming**

Stemming is the process of reducing words to their simplest form. This would allow us to group terms that are similar together. Poster Stemmer is utilized for implementation.

**4.5. Stop Words Removal**

Because they are of little use in training, the widely used words known as stop words need to be omitted and may also lead to incorrect results if not ignored. There is a collection of stop words in the NLTK library which can be used as a guide for removing stop words from the dataset. Stop words that are deleted from tweets using the Tf-Idf, Stanford, and wiki features.

**4.6. POS Tagger**

The tokenized text is allocated to the respective sections of the speech by using a POS tagger to improve the quality of the remaining data. Since other parts of speech are not of much importance, this can be used to remove only adjectives, nouns and adverbs. Example: 'I LOVE CODING' is extracted-'I 'is a pronoun, rest is omitted.

**4.7. Lemmatization**

It's used to apply stem words to their stems. Lemmatization refers to performing things correctly utilizing a vocabulary and morphological word analysis, mainly aimed at merely removing inflectional endings and restoring the root or dictionary form of a word known as the lemma, utilizing the Stanford core NLP kit.

A bag of words is generated after all of these pre-processing stages. The number of occurrences of each word is calculated in a bag of words, which is subsequently utilized as a feature to train a classifier.

**4.8. Bag of words or One Hot Encoding**

Each part of the vector corresponds to one word or n-gram (token) within the corpus vocabulary during this methodology. Then if the token at a selected index exists within the document, that part is marked as one, else it's marked as zero.

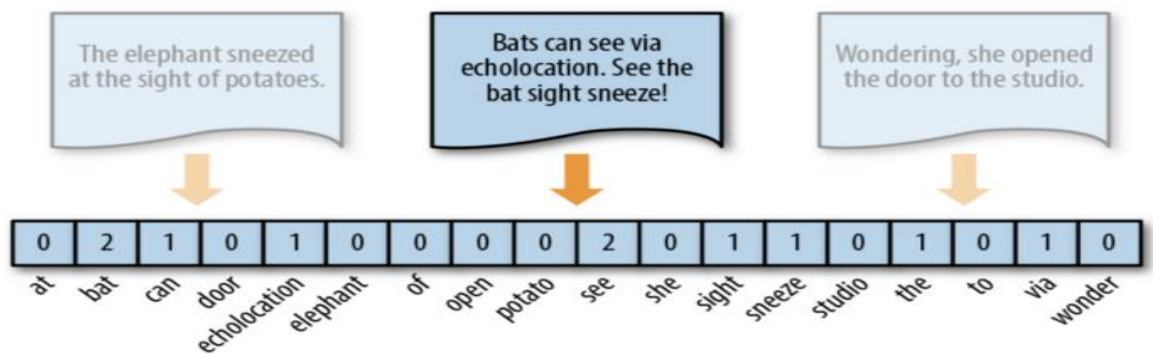

Fig.5: BOW Representation

In the above example, our corpus consists of every unique word across the three documents. A bag of Words model, or BOW for short, is a method of extracting text properties for use in modeling, such as machine learning techniques. The method is straightforward and adaptable, and it may be used to extract information from documents in a variety of ways. A bag of words is a text representation of the words that appear within a document.

It involves two ways:

1. A vocabulary of known words.

2. The measure of the presence of known words should be considered.

It's dubbed a bag of words because all information in the text about the sequence or arrangement of words is discarded. The model simply considers where recognized phrases appear in the document, not where they exist in the document. The histogram of the words inside the text is examined in this method. The bag of words approach (BOW), which considers each word count as a function, is a popular role extraction method for sentences and texts. The assumption is that documents are similar if they have comparable content. Furthermore, we can deduce something about the document's significance solely from its content. You can make the word bag as simple or as complex as you desire. The difficulty arises from deciding how to create the lexicon of knows words (or tokens) as well as how to rate the presence of such words.

### 4.9 Term Frequency-Inverse Document Frequency

TF-IDF stands for term frequency-inverse document frequency. It's a statistical metric for determining how important a word in a corpus of documents is to a text. This importance is related to the number of times a word appears in the text, but is offset by the number of documents in the corpus that contain that word.

### 4.10 Count Vectorization

The number of occurrences in a document for each word (i.e. independent text, such as an article, book, or simply a paragraph!) can be tallied using Count Vectorization. The Sci-kit learning library in Python offers a tool called CountVectorizer that can help with this. "The weather was nice today, so I went outdoors to enjoy the gorgeous and sunny weather," for example. The words "the," "weather," and "and" all appeared twice in the performance below, whereas other words only appeared once. Count Vectorization achieves this goal.

### 4.11 Distributional Similarity Based Representations- Word Embedding's

A word embedding may be a sort of learnt text illustration (real valued vectors) during which words with connected meanings square measure drawn equally, like the favored "King-Man + girl = Queen" example. For every word, the idea of employing a dense, distributed illustration is central to the approach. Every term may be a real-value vector with tens or many dimensions. In distinction, distributed word representations, like one-hot encryption, need thousands or countless dimensions. The 2 most well liked word embedding's square measure Word2Vec and Glove:

**4.12. Word2Vec:**

There square measure primarily 2 versions of Word2Vec — Skip-Gram and Continuous Bag of Words (CBOW). The embedding is learned by the CBOW model anticipating the present word supported its context (surrounding words). Given a current word, the Skip-Gram model learns by predicting the encompassing words (context). Word2vec represents every separate word with a selected list of numbers referred to as a vector, because the name suggests. The vectors square measure a basic function (the similarity of the trigonometric function between the vectors) is employed. Indicates the amount of linguistics similarity between the words drawn by those vectors. Skip-gram seeks to predict a given word's immediate neighbors. We tend to take a central word and a context window (neighbors) terms, and take a few window step sizes round the central word, we tend to attempt to predict the context words. So, our model can describe a distribution of chance, i.e. the chance of a word occurring in context given a central word, and to maximize the chance, we'll select our vector representations. We tend to take away the output layer and use the hidden layer to urge our word vectors till we are able to predict the encompassing words with an affordable degree of exactitude. We tend to begin with little random data formatting of vectors of terms. By minimizing the loss perform, our prognosticative model learns the vectors. In Word2vec, this happens with a feed-forward neural network with a language modeling task (predict next word) and improvement techniques like random gradient descent. This measures the options that I actually have used for machine learning algorithms like SVM, Naïve Thomas Bayes, and Random Forest.

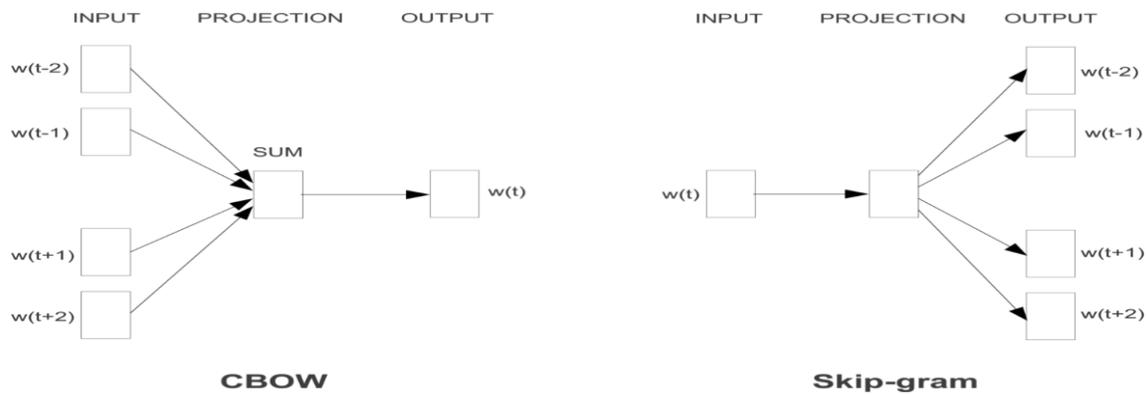

Fig. 6: Word2Vec CBOW vs Skip-gram

**4.13. Gensim Model for DOC2VEC used in LSTM**

Gensim is paid as a kit of natural language processing that 'Topic Modeling for Humans' does. But it's even more than that, technically. I used Doc2Vec for LSTM, meaning the model of a deep learning algorithm, so clarify Doc2Vec first. I have already mentioned the Word2Vec I used in machine learning algorithms since I used algorithms for machine learning as well as deep learning algorithms. So, for machine learning, I used Word2Vec and I used Doc2Vec for the LSTM model. The aim of doc2vec, regardless of its length, is to construct a numeric representation of a text, as mentioned. But records do not come in logical forms like words and unlike words, so another technique needs to be found. Because the Doc2vec model is an unsupervised process, it should be slightly adjusted to "participate" in this challenge. Fortunately, as in most circumstances, we can utilize a couple tricks: remember how we introduced another document vector in fig 3 that was unique to each document? If you think about it, you can add more vectors that don't have to be unique: for example, if we have tags for our documents (which we have), we can add them and receive their representation as vectors.

Additionally, they don't have to be unique. This way, we can add to the unique document tag one of our 17 tags, and create a doc2vec representation for them as well! See below:

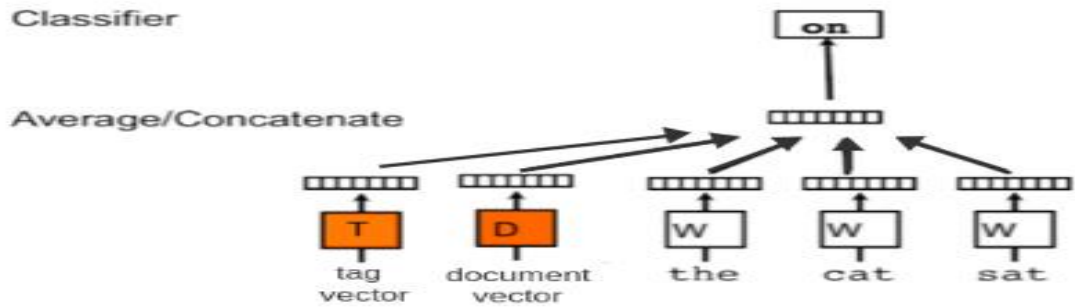

Figure. 8: Doc2Vec model with tag Vector

**4.14. Training**

From the training set both label and tweet are required by the classifier. The training set in this example refers to the set of tweets that must be processed before being fed into a classifier. The set of tweets must be transformed to a vector representation for further processing. The set of labels corresponding to each tweet is also provided into the classifier as a vector.

**4.15. Saving the classifier and the count vectorizer object**

Since training needs to be done once, it is necessary to load the trained classifier object into a pickle file. Same is applicable with the count Vectorizer Object. It count the number of words present in the pickle file using the TF-IDF method. And then the testing phase takes place for unseen data.

**4.16. Testing**

**Loading of saved models**: Qualified classification models are loaded from the pickle file to be used for test dataset prediction (Individual tweet profiling data). The test dataset is preprocessed in a way that is similar to the data from the training for test tweets class prediction. Each tweet is graded into a class that is depressed or non-depressed. We calculate the confusion matrix for the assessment of classification results based on the values of true or false positives and negatives. Confusion Matrix is used for how much our classifier shows accurate results as well as false results.

**V. RESULTS**

The depression detection system that we designed uses emotional artificial intelligence for the detection of depression using online social profiles. We used supervised learning models and Deep Learning algorithm i.e Long Short Term Memory (LSTM) were training of the classifier was given using the emotion dataset. The dataset has 15,000 tweets manually annotated by us. The highest accuracy of the classifier is around 70% for LSTM and for SVM the highest accuracy is 81.79%. I have trained the classifier on the datasets that are widely used in literature for the emotion mining task. The results are presented using the classification metrics like accuracy and confusion matrix.

**Confusion Matrix:** A confusion matrix is a machine learning classification performance evaluation methodology. It's a type of table that allows you to see how well a classification model performs on a set of test data for which the true values are known. True Positive (TP), True Negative (TN), False Positive (FP), False Negative (FN) these given four parameters of the confusion matrix are used to calculate the Accuracy, Recall, Precision and FI-Score, with the help of these four classes we compute all values. The results are evaluated on the basis of F1 score and accuracy. The F1 score is the primary performance measure and accuracy is the secondary measure with the help of confusion matrix we call Precision, Recall, F1 Score and Accuracy it simply shows how much our classifier shows accurate results or with the help of given formulas:

**A. Precision**: TP/TP+FP, where TP is true positive, FP is false positive.

**B. Recall**: TP/TP+FN.

**C. FI Score:** 2* (Precision*Recall)/ Precision + Recall

The dataset was balanced having 7,500 tweets in each data class that is depressive and non-depressive. SVM classifier its accuracy is more than other classifiers. We choose SVM classifier for unseen data. We trained our system using LSTM (Deep Learning Classifier), SVM, Naïve Bayes and Random Forest classifier (Machine Learning Classifier). The classifiers accuracy is described in the below table:

| Classifier | Precision | Recall | F1 Score | Accuracy |
|---|---|---|---|---|
| LSTM | 0.88 | 0.60 | 0.71 | 70.00% |
| Multinomial NB | 0.79 | 0.78 | 0.78 | 79.32% |
| SVM | 0.79 | 0.82 | 0.79 | 81.79% |
| Random Forest | 0.75 | 0.82 | 0.78 | 80.37% |

**Table 1 Results of different Classifier performance**

Figure 9, 10, 11 and 12 show the performance of the system using confusion matrices:

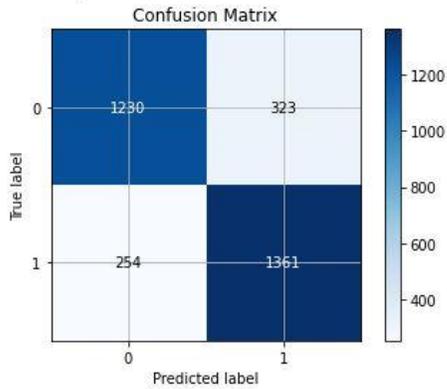
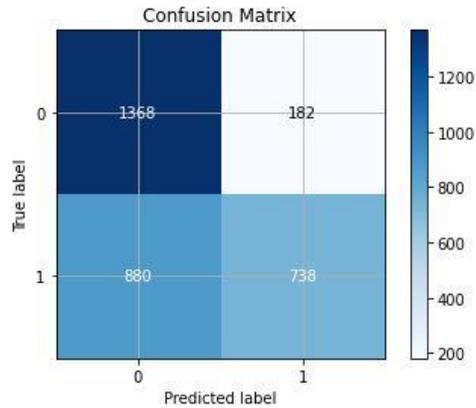

Fig. 9: Confusion Matrix of SVM          Fig.10: Confusion Matrix of LSTM

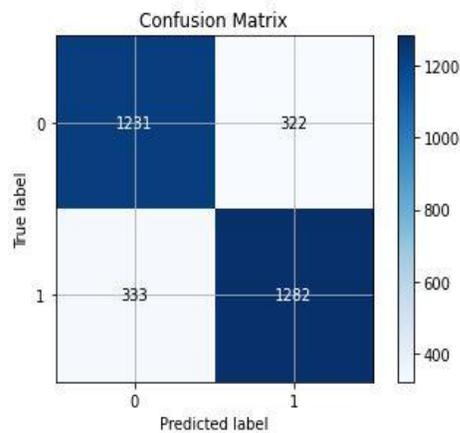
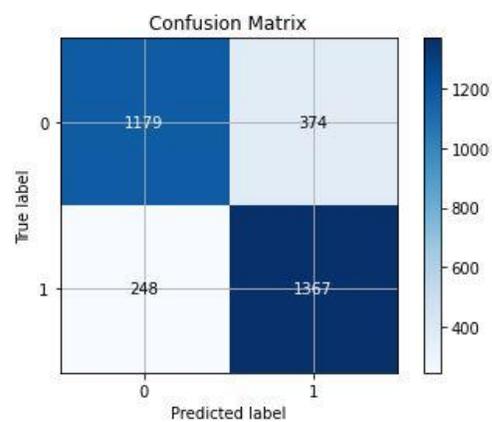

Fig.10: Multinomial Naïve Bayes          Fig. 12: Random Forest

## V1. CONCLUSION AND FUTURE WORK

The analytics performed on the chosen dataset offer some insight into the analysis issues. The subsequent may be an outline of our findings: What is depression, and what area unit the foremost current causes of depression? whereas we have a tendency to all feel irritable, unhappy, or low from time to time, some individuals experience these feelings on an everyday basis, over long periods of your time (weeks, months, or perhaps years), and in some cases for no obvious cause. Despondence is over simply a nasty mood; it is a state of affairs that has a bearing on an individuals physical and mental well-being. Any one can be suffering from depression at any time. However, bound stages or circumstances in modern day world expose people to anxiety and depression. During entire life the changes around or inside an individual all end in a surge of emotions which will contribute to unhappiness in introvert and socially separated people. We conducted a comparison of state-of-art deep learning models to pre-detect depression from tweets at the user level. We ran our models on a manually preprocessed dataset and discovered that SVM created higher results. In the future, we can use a special methodology to extract paraphrases from a wider set of emotional qualities. We will additionally check the quality and potency of our models with lot more datasets.